# Export Behaviour Modeling Using EvoNF Approach

Ron Edwards[1], Ajith Abraham[2] and Sonja Petrovic-Lazarevic[3]

[1]Monash University, School of Business and Economics,
McMahons Road, Frankston 3199, Australia
`ron.edwards@buseco.monash.edu.au`
[2]Department of Computer Science, Oklahoma State University,
700 N Greenwood Avenue, Tulsa, OK 74106-0700, USA
`ajith.abraham@ieee.org`
[3]Monash University, Department of Management,
McMahons Road, Frankston 3199, Australia
`sonja.petrovic-lazarevic@buseco.monash.edu.au`

**Abstract.** The academic literature suggests that the extent of exporting by multinational corporation subsidiaries (MCS) depends on their product manufactured, resources, tax protection, customers and markets, involvement strategy, financial independence and suppliers' relationship with a multinational corporation (MNC). The aim of this paper is to model the complex export pattern behaviour using a Takagi-Sugeno fuzzy inference system in order to determine the actual volume of MCS export output (sales exported). The proposed fuzzy inference system (FIS) is optimised by using neural network learning and evolutionary computation. Empirical results clearly show that the proposed approach could model the export behaviour reasonable well compared to a direct neural network approach.

## 1. Introduction

Malaysia has been pursuing an economic strategy of export-led industrialisation [3][6][7]. To facilitate this strategy, foreign investment is courted through the creation of attractive incentive packages. These primarily entail taxation allowances and more liberal ownership rights for investments [8][10][11]. The quest to attract foreign direct investment (FDI) has proved to be highly successful [9]. The bulk of investment has gone into export-oriented manufacturing industries.

Several specific subsidiary features identified in international business literature are particularly relevant when seeking to explain MNC subsidiary export behaviour. The location factors in attracting FDI to the country, the subsidiary's functional roles, size and age, and whether subsidiary products are targeted at niche or broader markets, have all been perceived to be determinants of export behaviour. This paper is concerned with the manner in which the structure and strategy of MNC that have invested in Malaysia affect the export intensity of their subsidiaries. Prior to going into the details of the study, it is important to explain that there are two related aspects of export behaviour. One aspect is the probability of a firm exporting at all. The other aspect is the relationship between the percentage of total sales exported and the size of the firm. According to the literature, larger firms are more likely to export. However, there is no clear relationship between size of the firm and export intensity. For example Bonnaccorsi [4] found that although larger firms were more likely to export, there was no significant difference between the export intensity of small, medium, or large firms. Wolff et al [12] also found no significant difference in export intensity between small, medium and large firms. They argued that the type of resources available is a key factor, specifically, that with the appropriate type of resource, a small firm can use the same competitive patterns utilised by larger firms with the same effectiveness. Wagner (2001) notes that greater firm size is neither necessary nor sufficient for any industry or country.

In this paper we are concentrated on the MCS product manufactured, resources, tax protection, customers and markets, involvement strategy, financial independence and suppliers' relationship with a MNC. We use the EvoNF, an integrated computational framework to optimise FIS through neural network learning and evolutionary computation [1].

The paper is divided as follows: Section 2 explains the role of fuzzy inference systems in determining the export behaviour of MCS. Section 3 illustrates the experimentation results based on data provided by Malaysian MCS. The paper ends with concluding remarks.

## 2. The Role of FIS for Explaining the Export Behaviour of MCS

A FIS can utilize human expertise by storing its essential components in rule base and database, and perform fuzzy reasoning to infer the overall output value. The derivation of *if-then* rules and corresponding membership functions (MF) depends heavily on the researcher's *a priori* knowledge about the system under consideration. However, there is no systematic way of transforming experiences of knowledge of human experts to the knowledge base of a FIS. There is also a need for adaptability or some learning algorithms to produce outputs within the required error rate [2].

In this section, we define the architecture of EvoNF, as an integrated computational framework to optimise FIS by using neural network learning technique and evolutionary computation. The proposed framework could adapt to Mamdani, Takagi-Sugeno or other FIS. The architecture and the evolving mechanism can be considered as general framework for adaptive fuzzy systems. That is a FIS can change their MF (quantity and shape), rule base (architecture), fuzzy operators and learning parameters according to different environments without human intervention.

Solving multi-objective problems is, generally, a very difficult goal. In optimisation problems, the objectives often conflict across a high-dimension problem space and may also require extensive computational resources. The hierarchical evolutionary search framework could adapt MF (shape and quantity), rule base (architecture), fuzzy inference mechanism (T-norm and T-conorm operators) and the learning parameters of neural network learning algorithm. In addition to the evolutionary learning (global search) neural network learning could be considered as a local search technique to optimise the parameters of the rule antecedent/consequent parameters and the parameterised fuzzy operators.

Figure 1 illustrates the interaction of various evolutionary search procedures. For every type of FIS (for example Mamdani type), there exist a global search of learning algorithm parameter, inference mechanism, rule base and MF in an environment decided by the problem. Thus the evolution of FIS will evolve at the slowest time scale while the evolution of the quantity and type of MF will evolve at the fastest rate. The function of the other layers could be derived similarly.

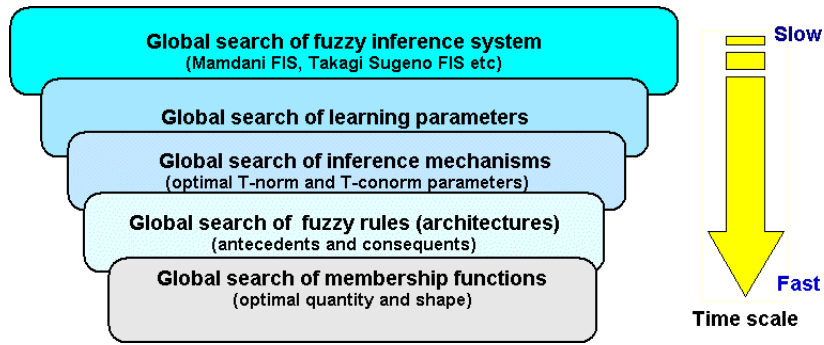

**Figure 1.** General computational framework for EvoNF

Hierarchy of the different adaptation layers (procedures) will rely on the prior knowledge. For example, if there is more prior knowledge about the architecture than the inference mechanism then it is better to implement the architecture at a higher level. If we know that a particular FIS will suit best for the problem, we could also minimize the search space. For fine-tuning the FIS all the node functions are to be parameterised.

### 2.1 Parameterization of Membership Functions

FIS is completely characterized by its MF For example, a generalized bell MF is specified by three parameters ($p, q, r$) and is given by:

$$\text{Bell}(x, p, q, r) = \frac{1}{1 + \left| \frac{x - r}{p} \right|^{2q}}$$

Figure 2 shows the effects of changing $p$, $q$ and $r$ in a bell MF. Similar parameterisation can be done with most of the other MF.

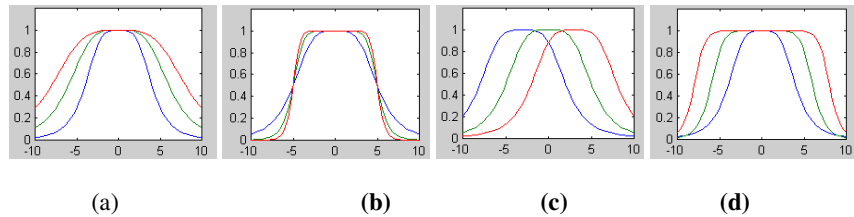

(a)  (**b**)  (**c**)  (**d**)

**Figure 2.** (**a**) Changing parameter $p$ (**b**) changing parameter $q$ (**c**) changing parameter $r$ (**d**) changing $p$ and $q$

## 2.2 Parameterization of T-norm operators

T-norm is a fuzzy intersection operator, which aggregates the intersection of two fuzzy sets A and B. The Schweizer and Sklar's T-norm operator can be expressed as:

$$T(a,b,p) = \left[\max\left\{0,(a^{-p}+b^{-p}-1)\right\}\right]^{-\frac{1}{p}}$$

It is observed that

$$\lim_{p \to 0} T(a,b,p) = ab$$

$$\lim_{p \to \infty} T(a.b,p) = \min\{a,b\}$$

which correspond to two of the most frequently used T-norms in combining the membership values on the premise part of a fuzzy *if-then* rule.

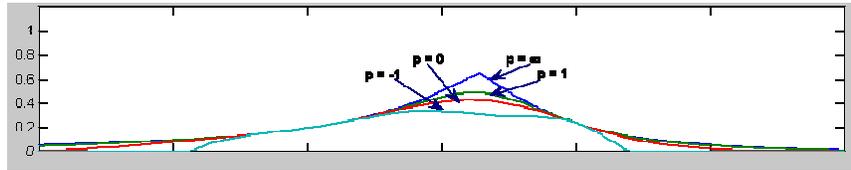

**Figure 3.** Effects of changing *p* of T-norm operator for two Bell MF

To give a general idea of how the parameter *p* affects the T-norm operator, Figure 3 illustrates T-norm operator $T_{(a,b,p)}$ for different values of *p*.

## 2.3 Chromosome Modeling and Representation

The antecedent of a fuzzy rule defines a local region, while the consequent the behaviour within the region via various constituents. Basically the antecedent part remains the same regardless of the inference system used. Different consequent describes constituents result in different FIS. For applying evolutionary algorithms, problem representation (chromosome) is very important as it directly affects the proposed algorithm. Referring to Figure 1 each layer (from fastest to slowest) of the hierarchical evolutionary search process has to be represented in a chromosome for successful modeling of EvoNF. A typical chromosome of the EvoNF would appear as shown in Figure 5 and the detailed modeling process is as follows.

**Layer 1:** The simplest way is to encode the number of MF per input variable and the parameters of the MF. Figure 5 depicts the chromosome representation of *n bell* MF specified by its parameters *p*, *q* and *r*. The optimal parameters of the MF located by the evolutionary algorithm will be later fine tuned by the neural network-learning algorithm. Similar strategy could be used for the output MF in the case of a Mamdani FIS. Experts may be consulted to estimate the MF shape forming parameters to estimate the search space of the MF parameters.

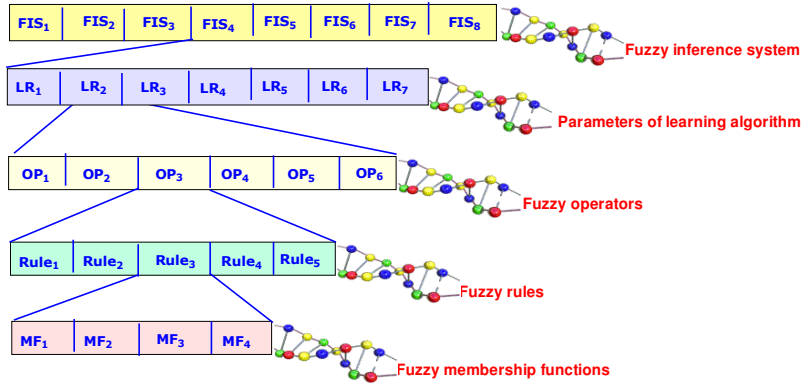

**Figure 4.** Chromosome structure of the EvoNF model

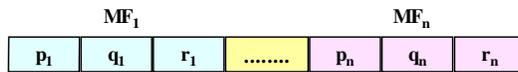

**Figure 5.** Chromosome representing *n* MF for every input/output variable coding the parameters of a bell shape MF

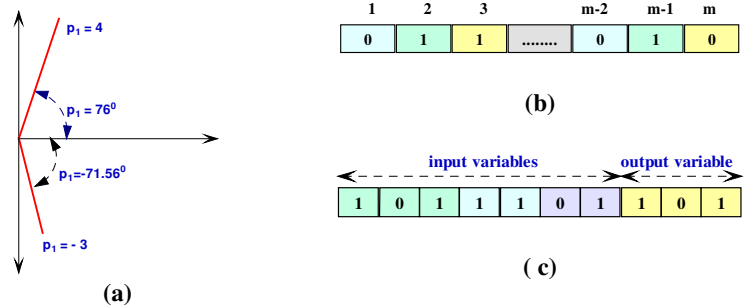

**Figure 6.** (a) Angular coding technique (b) representation of *m* fuzzy rules (c) representation of I/O variables

We used the angular coding method proposed by [5] for representing the rule consequent parameters of the Takagi-Sugeno inference system. Rather than directly coding the consequent parameters, the "transformed" parameters represent the direction of the tangent $\alpha_i = \arctan p_i$. The range for the parameters $\alpha_i$ is the interval ($-90^0$, $+90^0$), such that the parameters $p_i$ can assume any real value. A single input Takagi-Sugeno system $Y = p_1 X + p_0$ defines a straight line. The real value $p_1$ is simply the gradient between this line and the *X*-axis. Parameter $p_0$ determines the offset of the straight line (intercept) along the *Y*-axis. The procedure is illustrated in Figure 6.

**Layer 2.** This layer is responsible for the optimisation of the rule base. This includes deciding the total number of rules, representation of the antecedent and consequent parts. The simplest way is that each gene represents one rule, and "1" stands for a selected and "0" for a non-selected rule. Figure 6 (b) displays such a chromosome structure representation. To represent a single rule a position dependent code with as many elements as the number of variables of the system is used. Each element is a binary string with a bit per fuzzy set in the fuzzy partition of the variable, meaning the absence or presence of the corresponding linguistic label in the rule. For a three input and one output variable, with fuzzy partitions composed of 3,2,2 fuzzy sets for input variables and 3 fuzzy sets for output variable, the fuzzy rule will have a representation as shown in Figure 6(c).

**Layer 3.** In this layer, a chromosome represents the different parameters of the T-norm and T-conorm operators. Real number representation is adequate to represent the fuzzy operator parameters. The parameters of the operators could be fine-tuned using gradient descent techniques.

**Layer 4.** This layer is responsible for the selection of optimal learning parameters. Performance of the gradient descent algorithm directly depends on the learning rate according to the error surface. The optimal learning parameters decided by the evolutionary algorithm will be used to tune MF and the inference mechanism.

**Layer 5.** This layer basically interacts with the environment and decides which FIS (Mamdani type and its variants, Takagi-Sugeno type, Tsukamoto type etc.) will be the optimal according to the environment. Once the chromosome representation, $C$, of the entire EvoNF model is done, the evolutionary search procedure could be initiated as follows:

1. *Generate an initial population of N numbers of C chromosomes. Evaluate the fitness of each chromosome depending on the problem.*
2. *Depending on the fitness and using suitable selection methods reproduce a number of children for each individual in the current generation.*
3. *Apply genetic operators to each child individual generated above and obtain the next generation.*
4. *Check whether the current model has achieved the required error rate or the specified number of generations has been reached. Go to Step 2.*
5. *End*

## 3. Model Evaluation and Experimentation Results

For simulations we have used data provided from a survey of 69 Malaysian MCS. Each corporation subsidiary data set were represented by the following input variables:
- Product manufactured (1 -5 scale representing fully independent from the parent and fully dependent)
- Resources (1 - 5 scale representing fully independent from the parent and fully dependent)

- Tax protection (1 - 5 scale representing tax protection and no tax protection)
  Customers and market (1 - 4 scale representing the geographical distribution of the customers)
- Involvement strategy (1 - 4 scale representing subsidiary, subsidiary and parent, parent alone and equal share)
- Financial independence (1-5 scale representing fully independent from the parent and fully dependent)
- Suppliers relationship (1 - 5 scale representing fully independent from the parent and fully dependent)

### 3.1 EvoNF training

We used the popular grid partitioning method (clustering) to generate the initial rule base. This partition strategy requires only a small number of MF for each input. We used the 90% of the data for training and remaining 10% for testing and validation purposes. The initial populations were randomly created based on the parameters shown in Table 1. We used a special mutation operator, which decreases the mutation rate as the algorithm greedily proceeds in the search space 0. If the allelic value $x_i$ of the $i$-th gene ranges over the domain $a_i$ and $b_i$ the mutated gene $x_i'$ is drawn randomly uniformly from the interval $[a_i, b_i]$.

$$x_i' = \begin{cases} x_i + \Delta(t, b_i - x_i), \text{if } \omega = 0 \\ x_i + \Delta(t, x_i - a_i), \text{if } \omega = 1 \end{cases}$$

where $\omega$ represents an unbiased coin flip $p(\omega = 0) = p(\omega = 1) = 0.5$, and

$$\Delta(t, x) = x \left(1 - \gamma^{\left(1 - \frac{t}{t_{max}}\right)^b}\right)$$

defines the mutation step, where $\gamma$ is the random number from the interval $[0,1]$ and $t$ is the current generation and $t_{max}$ is the maximum number of generations. The function $\Delta$ computes a value in the range $[0,x]$ such that the probability of returning a number close to zero increases as the algorithm proceeds with the search. The parameter $b$ determines the impact of time on the probability distribution $\Delta$ over $[0,x]$. Large values of $b$ decrease the likelihood of large mutations in a small number of generations. The parameters mentioned in Table 1 were decided after a few trial and error approaches. Experiments were repeated 3 times and the average performance measures are reported. Figures 10 illustrates the meta-learning approach for training and test data combining evolutionary learning and gradient descent technique during the 35 generations.

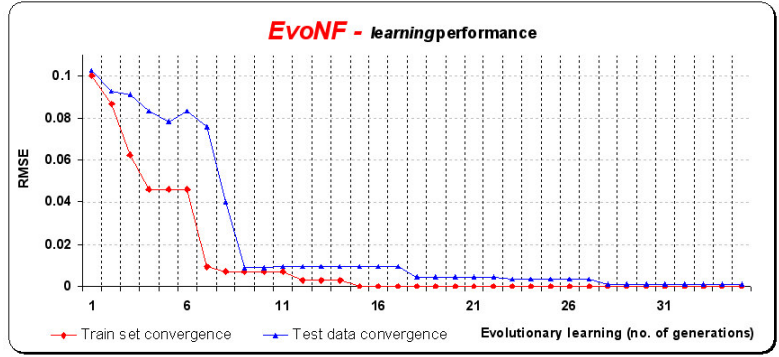

**Figure 7.** Meta-learning performance (training and test) of EvoNF framework

The 35 generations of meta-learning approach created 76 *if-then* Takagi-Sugeno type fuzzy *if-then* rules compared to 128 rules using the grid-partitioning method. We also used a feed forward neural network with 12 hidden neurons (single hidden layer) to model the export output for the given input variables. The learning rate and momentum were set at 0.05 and 0.2 respectively and the network was trained for 10,000 epochs using BP [2]. The network parameters were decided after a trial and error approach. The obtained training and test results are depicted in Table 2 (CC=correlation coefficient).

**Table 1.** Parameter settings of EvoNF framework

| Population size | 40 |
|---|---|
| Maximum no of generations | 35 |
| FIS | Takagi Sugeno |
| Rule antecedent MF | 2 MF (parameterised Gaussian)/ Input |
| Rule consequent parameters | Linear parameters |
| Gradient descent learning | 10 epochs |
| Ranked based selection | 0.50 |
| Elitism | 5 % |
| Starting mutation rate | 0.50 |

**Table 2.** Training and test performance of the different intelligent paradigms

| | Intelligent paradigms | | | | | |
|---|---|---|---|---|---|---|
| | EvoNF | | | Neural network | | |
| Export output | RMSE | | *CC | RMSE | | *CC |
| | Train | Test | | Train | Test | |
| | 0.0013 | 0.012 | 0.989 | 0.0107 | 0.1261 | 0.946 |

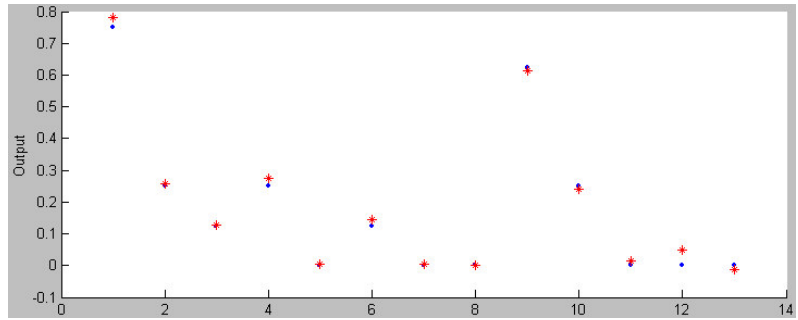

**Figure 8**. Test results showing the export output (scaled values) for 13 MNC's with respect to the desired values (*).

## 4. Conclusions

Our analysis on the export behavior of Malaysia's MCS reveals that the developed EvoNF model could learn the chaotic patterns and model the behavior using an optimized Takagi Sugeno FIS. As illustrated in Figure 8 and Table 2, EvoNF could easily approximate the export behavior within the tolerance limits. When compared to a neural network approach, EvoNF performed better (in terms of lowest RMSE) and better correlation coefficient. Our experiment results also reveal the importance of all the key input variables to model the behavior within the required accuracy limits. These techniques might be useful not only to MNC's but also to administrators and Government for long-term strategic management of the economy.

As a future research, we also plan to incorporate more intelligent paradigms to improve the modeling aspects of the export behavior.